\documentclass[11pt,a4paper]{article}
\usepackage[hyperref]{AACL-IJCNLP2020}
\usepackage{times}
\usepackage{latexsym}

\usepackage{microtype}

\aclfinalcopy 

\usepackage{multirow}
\usepackage{dirtytalk}
\usepackage{rotating}
\usepackage{arydshln}
\usepackage{xcolor}

\title{Using Multiple Subwords to Improve English-Esperanto Automated Literary Translation Quality}

\author{ Alberto Poncelas\textsuperscript{1,2}, Jan Buts\textsuperscript{1,2}, James Hadley\textsuperscript{1,2}, Andy Way\textsuperscript{2} \\
 \textsuperscript{1}Trinity Centre for Literary and Cultural Translation, Trinity College Dublin, Ireland \\
  \texttt{butsj@tcd.ie,hadleyj@tcd.ie} 
  \\
  \textsuperscript{2}ADAPT Centre, School of Computing, Dublin City University, Ireland \\
  \texttt{name.surname@adaptcentre.ie} 
  }

\date{}

\begin{document}
\maketitle
\begin{abstract}

Building Machine Translation (MT) systems for low-resource languages remains challenging. For many language pairs, parallel data are not widely available, and in such cases MT models do not achieve results comparable to those seen with high-resource languages. 

When data are scarce, it is of paramount importance to make optimal use of the limited material available. To that end, in this paper we propose employing the same parallel sentences multiple times, only changing the way the words are split each time. For this purpose we use several Byte Pair Encoding models, with various merge operations used in their configuration. 

In our experiments, we use this technique to expand the available data and improve an MT system involving a low-resource language pair, namely English-Esperanto. 

As an additional contribution, we made available a set of English-Esperanto parallel data in the literary domain.

\end{abstract}
\section{Introduction}

In this paper, we use the constructed language Esperanto to illustrate potential improvements in the automatic translation of material from low-resource languages. Languages are considered low-resource when there is little textual material available in the form of electronically stored corpora. They pose significant challenges in the field of Machine Translation (MT), since it is difficult to build models that perform adequately using small amounts of data. 

Multiple techniques have been developed to improve MT in conditions of data scarcity. A popular approach is to translate indirectly via a pivot language~\citep{utiyama2007comparison,firat2017multi,liu2018pivot,poncelas2020impact}. Moreover, indirect translation can be used for creating additional training data. A further useful technique for expanding the dataset is back-translation~\citep{sennrich2016}. This procedure consists of automatically translating a monolingual text from the target language into the selected source language, and then using the resulting parallel set as training data so the model benefits from this additional information. Although the quality of these sentence pairs is not as high as that of human-translated sentences (the source side contains mistakes produced by the MT system), the pairs are still useful when used as training data, because they do often improve the models~\citep{poncelas2019adaptation}.

Nonetheless, for some languages, the available data are in such short supply that MT models used for generating back-translated sentences may produce a high proportion of noisy sentences. The use of noisy sentences for building MT models could ultimately have a negative impact on the quality of the MT system's outputs~\citep{goutte2012impact}, and therefore they are often removed~\citep{khadivi2005automatic,taghipour2010discriminative,popovic2020extracting}.

We propose employing another technique to augment datasets: using the same set of sentences multiple times, but in slightly altered form each time. Specifically, we modify the sentences by using different Byte Pair Encoding (BPE)~\citep{sennrich2016neural} merge operations. We perform a fine-grained analysis, exploring the use of different splitting options on the source side, on the target side, and on both sides.

\section{Previous work}

This research is inspired by techniques for augmenting the training set artificially. One of these techniques is back-translation \citep{sennrich2016}, which involves creating artificial source-side sentences by translating a monolingual set in the target language. Similar techniques include the use of several models to generate sentences~\citep{poncelas2019combining,soto2020selecting}, or the use of synthetic data on the target side~\citep{chinea2017adapting,li2020revisiting}.

A technique that involves multiple segmentation is subword regularization \citep{kudo2018subword}, in which candidate sentences with different splits are sampled, either probabilistically or using a language model for training.

In the work of \citet{poncelas2020multiple}, different splits are used to build an English-Thai MT model. As the Thai language does not use whitespace separation between words, different splits can be applied, to address the fact that all the words and sub-words are joined together in the final output. 

More recently, \citet{provilkov2020bpe} introduced BPE-dropout, an improvement on standard BPE consisting of randomly dropping merges when training the model, such that a single word can have several segmentations.

\section{The Esperanto language}
This article is concerned with improving MT models for Esperanto, the most successful constructed international language~\citep{blanke2009causes}. It  was created in the late nineteenth century, and is said to be currently spoken by over 2 million people, spread across more than 100 countries \citep{ethnologue2020}. During its first century of development, Esperanto was principally maintained by means of membership-based organisations. Currently, internet applications such as Duolingo are supporting the wider spread of the language among new enthusiasts. While many Esperanto speakers have sought to develop the language through translation, the body of work available - particularly in digital formats - remains relatively small, making Esperanto a clear example of a low-resource language. 

Esperanto loosely derives its lexicon from several Indo-European languages, and shares some typological characteristics with, among others, Russian, English, and French \citep{parkvall2010typology}. In contrast to most natural languages, Esperanto's most distinctive characteristic is its regularity. The grammar consists of a very limited set of operations, to which there are, in principle, no exceptions. Furthermore, the language is agglutinative, and its suffixes are independently meaningful and invariable. For instance, \textit{virino}, the word for \say{woman}, consists of the compound parts \textit{vir} [adult human], \textit{in} [female], and \textit{o} [entity] (as the 'o' ending is used for all nouns). The word for \say{mother}, \textit{patrino}, largely refers to the same semantic categories, and is therefore structurally highly similar. 

As a consequence of this internal consistency, Esperanto learners can quickly expand their vocabulary by learning to segment words into their various parts, which can then be used to construct new words by morphological analogy. Because of its affinity with many other languages, and because of the thoroughly logical composition of its vocabulary, Esperanto has historically been central to several experiments in MT, most notably regarding its potential function as a pivot language between European languages~\citep{gobbo2015esperantomachine}. In this study, however, we focus on automatic translation into Esperanto for its own sake.

\section{Research Questions}

\begin{figure*}
\includegraphics[width=15.5cm, height=8.5cm]{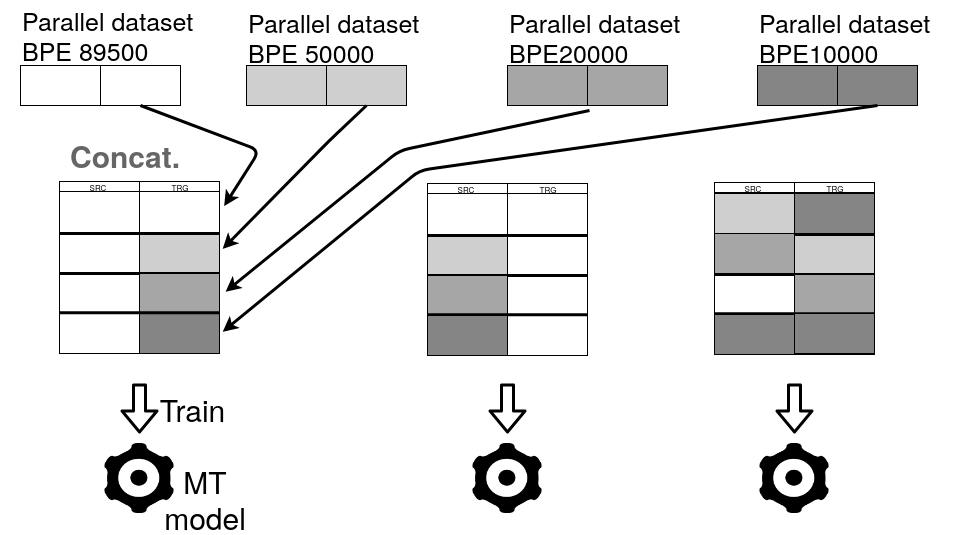}
\centering
\caption{Diagram with the experiments}
\label{fig:experiments}
\end{figure*}

We propose building MT models using training data composed of a dataset split into multiple variants with a different configuration of BPE, as presented in Figure \ref{fig:experiments}. At the top of the figure, one can see that the same parallel set has been processed using BPE with 89,500, 50,000 and 10,000 operations (trained separately for each language). The MT model represented on the left has been built using the same dataset replicated three times, the only difference being that on the target side, different splits were implemented. Similarly, the MT model in the centre is built with different splits on the source side. The last model, represented on the right, combines different splits both on the source and the target side.

In order to evaluate the models, we use a test set that is split with a single BPE strategy (i.e. using 89,500 merge operations, the default proposed in the work of~\citet{sennrich2016neural}). Therefore, using different merge operations on the source side of the training data may not have as big an impact as when they are applied to the target side (not all the words will match those in the test set). However, the addition of other BPE configurations could in principle still be useful to improve modeling for the source language.

In Section \ref{sec:settings} we describe the settings of the MT and the data used for training. In Section \ref{sec:baseline_MT} we analyze the results achieved by the baseline system.

This paper's experiments are divided into three sections. Each of these sections describes and also provides the evaluation of a model. The sections are the following:

\begin{itemize}
    \item Combination of dataset with different merge operations on the target side (Section \ref{sec:exp_target}).
    \item Combination of dataset with different merge operations on the source side (Section \ref{sec:exp_source}).
    \item Combination of dataset with different merge operations on both the source and target side (Section \ref{sec:exp_source_target}).
\end{itemize}

In Section \ref{sec:output_comparison}, we compare translation examples from the different models and analyze the different outcomes.

Finally, in Section \ref{sec:conclusion} we conclude and propose how these experiments could be expanded in future work.

\section{Experimental Settings}
\label{sec:settings}

The NMT systems we build are Transformer~\citep{vaswani2017attention} models, based on OpenNMT~\citep{opennmt}. Models are trained for a maximum of $30$K steps using the recommended parameters.\footnote{\url{https://opennmt.net/OpenNMT-py/FAQ.html}} We have selected the model with the lowest perplexity on the development set.

\subsection{Dataset}

For training the models we use the \textit{Tatoeba}, \textit{GlobalVoices} and \textit{bible-uedin}~\citep{christodouloupoulos2015massively} datasets from OPUS project.\footnote{\url{http://opus.nlpl.eu/}} Our dataset thus contains material from the Bible, from news sources, and from less domain-specific multilingual translation examples. The sentences are randomly shuffled, after which 302,768 sentences are used as a training set and the other 1,000 as our dev set. All the sentences are tokenized and truecased.

BPE is applied using several merge operations. We use 89,500 operations as a starting point and explore other splits that produce smaller subword units (by using a lower number of merge operations). In our experiments we work with 50,000, 20,000 and 10,000 operations.

We also concatenate the dev set using the same configuration of BPE.

\subsection{Test Set}

In order to evaluate the quality of the models, two test sets are translated. The test sets are the same for all models. In addition to tokenization and truecase, we also use BPE with 89,500 merge operations. We do not use (or combine) other BPE configurations. The translations are evaluated using the BLEU~\citep{papineni2002bleu} metric.

The first test set is taken from the OPUS (\textit{Books}) dataset ~\citep{tiedemann2012parallel} (1562 sentences). Specifically, the test set consists of material from two texts available in English and in Esperanto translation, namely Carroll's \textit{Alice's Adventures in Wonderland}~\citep{carroll1865alice} and Poe's \textit{The Fall of the House of Usher}~\citep{poe1839fall}.\footnote{\url{https://farkastranslations.com/bilingual_books.php}}

The second test set (which contains 1256 sentences) consists of an English and an Esperanto version of Oscar Wilde's \textit{Salomé}~\citep{wilde1891salome},~\footnote{\url{https://en.wikisource.org/wiki/Salom\%C3\%A9} and \url{http://www.gutenberg.org/ebooks/63064}} a play originally written in French. As an additional contribution to this paper, we have made a set of aligned sentences from the texts available via OPUS.\footnote{\url{http://opus.nlpl.eu/Salome-v1.php}}~

Both test sets are in the literary domain, which is especially challenging \cite{toral2018level} for MT models. Not only do the test sets contain numerous personal names and uncommon vocabulary, they are also highly creative and, at times, experimental. For instance, in \textit{Alice's Adventures in Wonderland}, grammatical and lexical principles are often challenged on purpose to portray a character's individual traits (i.e. the Mock Turtle sings of \textit{Beau--ootiful soo--oop!}. In \textit{Salome}, characters regularly produce complex similes and metaphors to describe one another. The text is a variation on a religious theme, and heavily draws on Biblical imagery. While such material is highly challenging, the inclusion of Biblical matter in the training data may have a positive impact on the overall results.

\section{Baseline MT}
\label{sec:baseline_MT}

\begin{table}
\centering
\begin{center}
\begin{tabular}{ |p{2cm}|p{1.5cm}|p{1.5cm}|}
\hline
	\textbf{Traindata}	&	\textbf{Books}&	\textbf{Salome}		\\
\hline
TRG89500	&	6.72	&	18.03	\\
TRG50000	&	6.42	&	18.74	\\
TRG20000	&	6.89	&	18.49	\\
TRG10000	&	6.65	&	18.41	\\
\hline
\end{tabular}
\caption{ 
BLEU scores of the \textbf{Books} and	\textbf{Salome} test sets when translated using the Baseline MT.
}
\label{table:baseline_trg}
\end{center}
\end{table}

In Table \ref{table:baseline_trg} we present the models trained with the training data using different merge operations on the target side. 

The rows of the table correspond to the evaluation of the model, using the same data. The only difference is the number of BPE merge operations that have been used on the target side.

As the test set is split using 89,500 merge operations, it would not be beneficial to apply BPE with merge operations other than 89,500 on the source side. In fact, when using BPE with 50,000, 20,000 and 10,000 operations on the source side, the BLEU score for the translation of the Books data is only 5.75, 5.70 and 5.76, respectively, and 14.30, 13.24, and 14.53 for \textit{Salome}

Table \ref{table:baseline_trg} shows that the four models achieve similar results. As mentioned before, the \textit{Books} set contains complex grammatical and lexical constructions, which makes it more difficult to translate. This is also evidenced in the table as BLEU scores of the \textit{Books} set are lower than those of the \textit{Salome} set. Moreover, there is no correlation between the number of merge operations and the performance. For example, we observe a small drop in the performance when decreasing the number of merge operations from 89,500 to 50,000, but the performance improves slightly when the number of operations is further decreased to 20,000. 

\section{Experiments}

\subsection{Different Merge Operations on the Target Side}
\label{sec:exp_target}

In the first set of experiments we explore the models when the sentences in the parallel set are replicated by changing only the number of BPE merge operations used on the target side. We perform two sets of experiments: one where we keep the duplicates (sentences that remain the same after being split with different BPE configurations), and another where duplicates are removed.

\begin{table*}
\centering
\begin{center}
\begin{tabular}{ |p{0.2cm}|p{8cm}|p{1cm}|p{1cm}|}
\hline
& Traindata	&	Books	&	Salome	\\
\hline
\multirow{10}{*}{\rotatebox[origin=c]{90}{ keep duplicates}}
& TRG89500 \&  TRG50000	&	6.60	&	18.25	\\
& TRG89500 \&  TRG20000	&	6.89	&	18.49	\\
& TRG89500 \&  TRG10000	&	\bf7.79*	&	\bf19.41*	\\
& TRG50000 \&  TRG20000	&	\bf7.17	&	17.96	\\
& TRG50000 \&  TRG10000	&	\bf6.94	&	\bf18.91	\\
& TRG20000 \&  TRG10000	&	\bf7.48*	&	\bf19.53*	\\
\hdashline
& TRG89500 \&  TRG50000 \&  TRG20000	&	 6.64	&	18.63    \\
& TRG89500 \&  TRG50000 \&  TRG10000	&	 \bf7.38	&	18.64    \\
& TRG89500 \&  TRG20000 \&  TRG10000	&	 \bf7.91*	&	\bf19.29    \\
& TRG50000 \&  TRG20000 \&  TRG10000	&	 \bf7.51*	&	\bf18.86    \\
\hdashline
 &TRG89500 \&  TRG5000 \&  TRG20000 \&  TRG10000	&	 \bf7.25	&	18.68    \\
\hline
\hline
\multirow{10}{*}{\rotatebox[origin=c]{90}{ remove duplicates}}
& TRG89500 \&  TRG50000	&	\bf7.47*	&	\bf18.51	\\
& TRG89500 \&  TRG20000	&	\bf7.56*	&	\bf19.08	\\
& TRG89500 \&  TRG10000	&	\bf7.98*	&	\bf18.78	\\
& TRG50000 \&  TRG20000	&	\bf6.93	&	18.47	\\
& TRG50000 \&  TRG10000	&	\bf7.54*	&	\bf18.62	\\
& TRG20000 \&  TRG10000	&	\bf7.79*	&	18.32	\\
\hdashline
& TRG89500 \&  TRG50000 \&  TRG20000	&	 \bf7.22	&	18.40    \\
& TRG89500 \&  TRG50000 \&  TRG10000		&	 \bf7.53*	&	\bf18.99    \\
& TRG89500 \&  TRG20000 \&  TRG10000		&	 \bf7.90*	&	\bf18.76    \\
& TRG50000 \&  TRG20000 \&  TRG10000		&	 \bf7.39	&	\bf19.07    \\
\hdashline
&TRG89500 \&  TRG5000 \&  TRG20000 \&  TRG10000		&	 \bf7.75*	&	\bf18.84    \\
\hline
\end{tabular}
\caption{ 
Model performance using different merge operations on the target side.
}
\label{table:target_BPE}
\end{center}
\end{table*}

In Table \ref{table:target_BPE} we present the results of the models when trained with a different concatenation of datasets. The first column specifies the datasets used in the training. For example, the row \textit{TRG89500 \&  TRG50000} indicates that the training set used for building the MT model consists of sentences split using 89,500 and 50,000 merge operations, respectively

We mark in bold those scores that exceed 6.89 BLEU points, i.e. the maximum score achieved by the baseline models presented in Table \ref{table:baseline_trg}. The scores receive an asterisk when the improvements are statistically significant at p=0.01. Statistical significance has been computed using Bootstrap Resampling \cite{koehn04}.

In the table, we find that scores tend to be higher when duplicate sentence pairs are removed. By doing this the dataset is reduced by between 30\% and 45\%. In the second subtable, all the BLEU scores indicate improvements over the baseline, whereas in the first subtable some models, such as \textit{TRG89500 \&  TRG50000}, have a lower score.

The best performance is seen when for the multiple settings used, the number of merge operations differs greatly. For example, the highest scores are achieved when mixing 89,500 and 10,000 operations (i.e. the \textit{TRG89500 \&  TRG10000} rows in both subtables), the uppermost and the lowermost number of operations  used in the experiments. The same principle holds true for those models built by combining three or four datasets.

\subsection{Different Merge Operations on the Source Side}
\label{sec:exp_source}

\begin{table*}
\centering
\begin{center}
\begin{tabular}{ |p{0.2cm}|p{8cm}|p{1cm}|p{1cm}|}
\hline
&	Traindata	&	Books	&	Salome	\\
\hline
\multirow{6}{*}{\rotatebox[origin=c]{90}{ keep duplicates}}
&	SRC89500 \&  SRC50000	&	\bf7.19	&	\bf19.07	\\
&	SRC89500 \&  SRC20000	&	\bf7.06	&	18.46	\\
&	SRC89500 \&  SRC10000	&	\bf7.21	&	\bf19.28	\\
\hdashline
&	SRC89500 \&  SRC50000 \&  SRC20000	&	\bf7.25	&	\bf19.16	\\
&	SRC89500 \&  SRC50000 \&  SRC10000	&	\bf7.09	&	\bf19.15	\\
&	SRC89500 \&  SRC20000 \&  SRC10000	&	\bf6.96	&	\bf18.98	\\
\hdashline
&	SRC89500 \&  SRC50000 \&  SRC20000 \&  SRC10000	&	6.79	&	18.08	\\
\hline
\hline
\multirow{6}{*}{\rotatebox[origin=c]{90}{ remove duplicates}}
&	SRC89500 \&  SRC50000	&	6.44	&	\bf18.54	\\
&	SRC89500 \&  SRC20000	&	\bf7.05	&	\bf18.86	\\
&	SRC89500 \&  SRC10000	&	\bf7.43*	&	\bf19.35*	\\
\hdashline
&	SRC89500 \&  SRC50000 \&  SRC20000	&	\bf7.42*	&	\bf18.63	\\
&	SRC89500 \&  SRC50000 \&  SRC10000	&	6.66	&	\bf19.37	\\
&	SRC89500 \&  SRC20000 \&  SRC10000	&	\bf7.22	&	\bf19.40*	\\
\hdashline
&	SRC89500 \&  SRC50000 \&  SRC20000 \&  SRC10000	&	\bf7.31&	\bf19.08		\\
\hline
\end{tabular}
\caption{ 
Model performance using different merge operations on the source side.
}
\label{table:source_BPE}
\end{center}
\end{table*}

The next set of experiments explores the use of several merge operations on the source side. In this case, when combining the datasets, we ensure that the SRC89500 set is used, as the test set has been processed using 89,500 operations. We present the results in Table \ref{table:source_BPE}. Those scores that are higher than the baselines of Table \ref{table:baseline_trg} are marked in bold.

Our observations are similar to those obtained in Section \ref{sec:exp_target}. The best results are observed when the duplicate sentences are removed (between 25\% and 40\% of the sentences are removed) and the merge operation settings are the furthest apart (89,500 and 10,000).

Most of the models using several BPE configurations on the source side perform better than the baseline models. However, when compared to the experiments in the previous section (Table \ref{table:target_BPE}), the performance is lower.

\subsection{Different Merge Operations on both Source and Target Side}
\label{sec:exp_source_target}

\begin{table}
\centering
\begin{center}
\begin{tabular}{ |p{4cm}|p{1cm}|p{1cm}|}
\hline
	Traindata	&	Books	&	Salome	\\
\hline
\footnotesize SRC89500 \& SRC10000 \& TRG89500 \& TRG50000	&	\bf7.99*	&	17.78	\\
\hdashline
All	&	\bf8.11*	&	\bf19.70*	\\
\hline
\end{tabular}
\caption{ 
Model performance using different merge operations both on the source and target side.
}
\label{table:source_target_BPE}
\end{center}
\end{table}

The last set of experiments consists of building a model with data created using different splits both on the source and on the target side.

We perform experiments based on the outcomes observed in the previous section. Thus, two models are built. One combines the datasets split using BPE with 89,500 and 10,000 merge operations (both source and target side) and the other model, \textit{All},  combines the dataset with all the splits (i.e. 89,500, 50,000, 20,000 and 10,000).\footnote{Note that we use all possible combinations. For example, the training set of the \textit{All} model is built combining $4*4=16$ datasets.
} The duplicates are removed, as this approach showed the best results.

We present the translation quality of the test set using these models in Table \ref{table:source_target_BPE}. We see that the use of different splits on both the source and target sides tends to achieve the best results when compared both to baselines and to the experiments in the previous sections\footnote{We observed that the output tends to be more similar to the splits following the TRG89500 configuration.}.

\section{Comparison of Outputs}
\label{sec:output_comparison}

\begin{table*}
\centering
\begin{center}
\begin{tabular}{ |p{2.5cm}|p{12.5cm}|}
\hline
system	&	sentence	\\
\hline
source	&	said Alice, very much confused, ``I don't think--"

``Then you shouldn't talk," said the Hatter .\\
reference	&	Alicio , tre konfuzite, respondis... : ``mi ne pensas-" 

``se vi ne pensas, vi ne rajtas paroli," diris la Ĉapelisto .	\\
\footnotesize TRG89500 	&	Alico, tre konfuzita; mi ne pensas .

``\textbf{do vi ne} parolu," diris la Hater.
\\
\scriptsize SRC89500 \&  SRC10000	& Alico... ``\textbf{vi ne} parolu," diris la Hatar.	\\
\scriptsize TRG89500 \&  TRG10000	& ``\textbf{vi do ne} parolu," diris la Hatter. 	\\
\footnotesize All	&	``\textbf{vi devus} diri, " diris la Hater.\\
\hline
source	&	but she did not venture to say it out loud .	\\
reference	&	sed tion ŝi ne kuraĝis diri laŭte .	\\
\footnotesize TRG89500 	&	sed ŝi ne \textbf{diris} tion laŭte .	\\
\scriptsize SRC89500 \&  SRC10000	&	sed ŝi ne \textbf{decidis diri} tion laŭte.	\\
\scriptsize TRG89500 \&  TRG10000	&	sed ŝi ne \textbf{intence diri} tion.	\\
\footnotesize All	&	sed ŝi ne \textbf{entreprenis diri} tion laŭte.	\\
\hline
source	&	it is like a knot of serpents coiled round thy neck.	\\
reference	&	ili similas al fasko da nigraj serpentoj, kiun oni \^{j}etis ĉirkaŭ vian kolon .	\\
\footnotesize TRG89500 	&	ĝi similas \textbf{nodon de turmentoj }ĉirkaŭ via kolo.	\\
\scriptsize SRC89500 \&  SRC10000	&	via kolo estas kiel \textbf{nodo streĉata}.	\\
\scriptsize TRG89500 \&  TRG10000	&	estas kiel \textbf{nodo da serpento}j ĉirkaŭ via kolo.	\\
\footnotesize All	&	ĝi estas kiel \textbf{nodo de serpentoj} ĉirkaŭ via kolo.	\\
\hline 
source	&	it is like a pomegranate cut in twain with a knife of ivory.	\\
reference	&	kiel granato, tranĉita per ebura tranĉilo .	\\
\footnotesize TRG89500 	&	kiel granato, kiu \textbf{falas en du} pecojn de eburo .	\\
\scriptsize SRC89500 \&  SRC10000	&	kiel granato, kiu \textbf{tranĉiĝis en du} per tranĉilo de eburo.	\\
\scriptsize TRG89500 \&  TRG10000	&	ĝi similas al granato, \textbf{tranĉita en du} kun tranĉilo de eburo.	\\
\footnotesize All	&	kiel granato \textbf{eltranĉita en du} kun tranĉilo ebura.	\\
\hline
\end{tabular}
\caption{ 
Translation examples from the test set.
}
\label{table:sentence_examples}
\end{center}
\end{table*}

In Table \ref{table:sentence_examples}, we show some translation examples of the models that, as discussed in the previous sections, achieved the best performance. We mark in bold some important differences across the translations.

The first example, drawn from \textit{Alice in Wonderland}, contains a joke. Alice, who is collecting her thoughts, aims to voice her opinion, and starts out by saying \textit{I don't think...}. Before she can finish her sentence, however, the Mad Hatter interrupts her by stating that in that case, she should not speak. The human Esperanto translation makes this joke very explicit by repeating the emphasis on 'not thinking', whereas in English the transition is more subtle. Two of the systems, while differing in exact word order, succeed in reproducing the joke (TRG89500 and TRG89500 \&  TRG10000). In the other two models, either the crucial element \textit{do} [so], which realises the inference, is omitted, or the meaning is mistakenly changed to a positive imperative: \textit{vi devus diri} [you should say]. It can further be observed in the sentences that none of the systems translates the Hatter's name meaningfully. Either the name remains the same, or it is slightly altered from the original, in a seemingly random manner. Interestingly, Alice's name is adapted to Alico, which conforms to the rule that all Esperanto names end in\textit{ -o} (or, in some cases \textit{-a}), but the adaptation does not equal the human choice for \textit{Alicio}. 

The second example, also taken from \textit{Alice's Adventures in Wonderland}, is concerned with a particular fixed expression in the English language: \textit{venture to say}. The baseline system does not translate this mark of politeness, while the other models do provide varying translations (i.e. \textit{decidis}, \textit{sukcesis} and \textit{entrepenis}, which correspond to the past tenses of the verb \textit{to decide}, \textit{to succeed} and \textit{to undertake}). While none of them is completely correct (when compared to the human translation), all of them are fairly transparent in context, and foreground different aspects of meaning contained in the English \textit{venture}.  

With reference to the \textit{Salome} test set, we find in the entire translated text numerous small and relatively inconsequential vocabulary differences across systems (e.g. \textit{veston} or \textit{mantelon} for referring to a piece of clothing), as well as varying preferences for orthographically similar verb tenses (e.g \textit{lacigis} or \textit{lacigas}, past and present tense of the verb \textit{to tire} or \textit{wear out}). At times, the systems differ in their translation of multi-word units such as \textit{sacred person}, which is translated either as the literal \textit{sankta homo} or as the more interpretative \textit{sanktulo} [saint]. Overall, the systems perform well when translating the play's dense symbolism, as illustrated in Table \ref{table:sentence_examples}. 

The examples in the table are similes, which start with the explicit comparative phrase \textit{it is like}. In the first example, the baseline system does not manage to reproduce the reference to \textit{serpentoj} [snakes], although the mention of \textit{turmentoj} [afflictions] does offer an interesting metaphorical perspective. The system \textit{SRC89500 \& SRC10000} does not produce a correct translation, but those systems trained with different splits on the target side (i.e. the \textit{SRC10000 \& SRC89500} and \textit{All} systems) provide a remarkably good translation of the source. Similarly, in the last example included in the table, the baseline system fails to reproduce the meaning of the original (the knife falls apart instead of cutting the fruit), whereas all systems with multiple segmentation are successful in conveying a variant of the poetic image presented in the source text. 

In short, the examples in Table \ref{table:sentence_examples} indicate that a combination of different merge operations may improve results for translation into Esperanto, a language for which limited resources are available. In a number of cases, the systems succeed in translating highly uncommon constructions in the context of humorous and poetic literary discourse. 

\section{Conclusion and Future Work}
\label{sec:conclusion}

In this work, we have aimed to improve an English-Esperanto MT system by using multiple instances of the same sentence pair, split with different configurations of BPE.

In our experiments, the best performance tends to be achieved when splitting strategies are applied both on the source and target side, duplicate parallel sentences are removed, and the number of merge operations used are very different from each other. In our experiments, the best results are achieved when all the split-combinations are used on both sides.

Although the goal of these experiments is to find a technique to improve the MT models when the available data are very limited, this technique could also be applied in scenarios where data are abundant. It should be noted that Esperanto is perhaps a particularly suitable candidate for word-split methods, as the language's vocabulary consists of fixed chunks that are combined to form transparent compounds. However, the techniques applied here are in principle language-independent.

Finally, although we demonstrated that combining sentences with different merge operations improves the model, in this paper we could not determine the best configuration to use. Similarly, the test set that we used was processed using 89,500 merge operations. If the test set had been processed with a different BPE configuration the performance could have been different, especially when using models with different split configurations on the source side. Extensions of this work could involve finding an optimal configuration for achieving the best results, or testing the performance when combined with other word-splitting techniques.

\section*{Acknowledgments}

This research has been supported by the ADAPT Centre for Digital Content Technology which is funded under the SFI Research Centres Programme (Grant 13/RC/2106).

The QuantiQual Project, generously funded by the Irish Research Council’s COALESCE scheme (COALESCE/2019/117).

\bibliography{aacl-ijcnlp2020}
\bibliographystyle{acl_natbib}

\end{document}